\documentclass{article}

\usepackage{microtype}
\usepackage{graphicx}
\usepackage{subfigure}
\usepackage{booktabs} 

\usepackage{hyperref}

\usepackage[accepted]{icml2024}

\usepackage{amsmath}
\usepackage{amssymb}
\usepackage{mathtools}
\usepackage{amsthm}

\usepackage[capitalize,noabbrev]{cleveref}

\theoremstyle{plain}

\theoremstyle{definition}

\theoremstyle{remark}

\usepackage{multirow}
\usepackage{longtable}
\usepackage{listings}
\usepackage[normalem]{ulem}
\usepackage{tabularx}
\usepackage{array}
\newcolumntype{Y}{>{\centering\arraybackslash}X}
\usepackage{float}
\usepackage{graphicx}
\graphicspath{ {./images/} }
\usepackage{makecell}
\usepackage{adjustbox}
\usepackage{booktabs}
\usepackage{multicol}
\usepackage{cleveref}
\usepackage{xcolor}
\usepackage{colortbl}
\definecolor{gray}{RGB}{220,220,220}
\usepackage{pgf-pie}
\usepackage{subcaption}
\usepackage{tcolorbox}
\usepackage[utf8]{inputenc}
\usepackage{xcolor}

\usepackage{siunitx}

\newtcbox{\hlprimarytab}{on line, rounded corners, box align=base, colback=green!10,colframe=white,size=fbox,arc=3pt, before upper=\strut, top=-2pt, bottom=-4pt, left=-2pt, right=-2pt, boxrule=0pt}
\newtcbox{\hlsecondarytab}{on line, box align=base, colback=red!10,colframe=white,size=fbox,arc=3pt, before upper=\strut, top=-2pt, bottom=-4pt, left=-2pt, right=-2pt, boxrule=0pt}

\newcommand{\dashifted}{\raisebox{0.5\depth}{\tiny$\downarrow$}}
\newcommand{\uashifted}{\raisebox{0.5\depth}{\tiny$\uparrow$}}
\newcommand{\da}[1]{{\scriptsize\hlsecondarytab{\dashifted{#1}}}}
\newcommand{\ua}[1]{{\scriptsize\hlprimarytab{\uashifted{#1}}}}
\newcommand{\llama}{Llama 2 (7B)}
\newcommand{\code}{Code Llama (7B)}
\newcommand{\mistral}{Mistral (7B)}
\newcommand{\counter}{\texttt{Counter}}
\newcommand{\assert}{\texttt{assert}}

\usepackage[textsize=tiny]{todonotes}

\icmltitlerunning{VerityMath: Advancing Mathematical Reasoning by Self-Verification Through Unit Consistency}

\begin{document}

\twocolumn[
\icmltitle{VerityMath: Advancing Mathematical Reasoning by\\
            Self-Verification Through Unit Consistency}




\begin{icmlauthorlist}

\icmlauthor{Vernon Toh Yan Han}{sutd}
\icmlauthor{Ratish Puduppully}{i2r}
\icmlauthor{Nancy F. Chen}{i2r,cnrs,cfar}

\end{icmlauthorlist}


\icmlaffiliation{sutd}{Singapore University of Technology and Design}
\icmlaffiliation{i2r}{Institute for Infocomm Research (I$^2$R), A$^{*}$STAR, Singapore}
\icmlaffiliation{cnrs}{CNRS@CREATE, Singapore}
\icmlaffiliation{cfar}{Centre for Frontier AI Research (CFAR), A$^{*}$STAR, Singapore}

\icmlcorrespondingauthor{Vernon Toh Yan Han}{vernon\_toh@mymail.sutd.edu.sg}
\icmlcorrespondingauthor{Ratish Puduppully}{puduppully@i2r.a-star.edu.sg}

\icmlkeywords{Machine Learning, ICML}

\vskip 0.3in
]



\printAffiliationsAndNotice{$^{*}$Most of the work was done while the first author was an intern at A$^{*}$STAR.} 

\begin{abstract}
Large Language Models (LLMs), combined with program-based solving techniques, are increasingly demonstrating proficiency in mathematical reasoning. For example, closed-source models such as OpenAI GPT-4 and Claude show excellent results in solving math word problems. However, progress in math word problem-solving for open-source LLMs is limited, and the challenges these models face are not well-studied. In this paper, we study the performance of strong open-source LLMs, including \llama, \code, and \mistral{} on math word problems using program-based solving techniques. Specifically, we analyze the outputs of these models when applied to math word problems and identify a category of problems that pose a significant challenge, particularly those involving quantities spanning multiple units. To address this issue, we propose a systematic approach by defining the units for each quantity and ensuring the consistency of these units during mathematical operations. We developed \emph{Unit Consistency Programs} (UCPs), an annotated dataset of math word problems, each paired with programs containing unit specifications and unit verification routines. We fine-tuned \llama, \code, and \mistral{} models with UCPs to produce their \emph{VerityMath} variants. Our findings indicate that our approach, which incorporates unit consistency, currently slightly underperforms compared to an approach that does not. To understand the reasons behind this, we conduct an in-depth error analysis and suggest options for future improvements. Our code and dataset are available at \url{https://github.com/vernontoh/VerityMath}.
\end{abstract}

\section{Introduction}
The ability to reason during the process of thinking and decision-making is a fundamental aspect of human intelligence. Replicating this ability in machines has been an objective in the field of Natural Language Processing. Large language models (LLMs) \cite{openai2023gpt4, anil2023palm} mark significant progress toward this goal, demonstrating remarkable proficiency across a range of tasks, including mathematical reasoning \cite{zhou2023solving, zhao2023automatic, zheng2023progressivehint}. Specifically, methods like Program Aided Language Model (PAL) \cite{gao2023pal} as well as Program of Thoughts (PoT) \cite{chen2022program} have demonstrated improvements in LLMs' ability to solve complex mathematical problems. These methodologies empower LLMs to formulate programs as intermediate reasoning steps and delegate the execution of these steps to a Python interpreter, thereby enhancing computational accuracy. 

However, open-source LLMs like those referenced in \cite{touvron2023llama, rozière2023code, jiang2023mistral} demonstrate limited success in math reasoning tasks. For example, after fine-tuning on the GSM8K-PAL dataset provided by \citet{jie2023leveraging}, \mistral{} achieves just 70.4\% accuracy on GSM8K \cite{cobbe2021training} (Ref \Cref{tab:Main Results Table}). Our analysis of the fine-tuned \llama, \code{} and \mistral{} reveals challenges in solving math word problems with multi-unit quantities. These issues are more pronounced in multi-step reasoning, where early errors can lead to incorrect final solutions. Our study thus identifies specific challenges the model faces.

\begin{figure*}[t]
\centering
\includegraphics[width=0.98\textwidth]{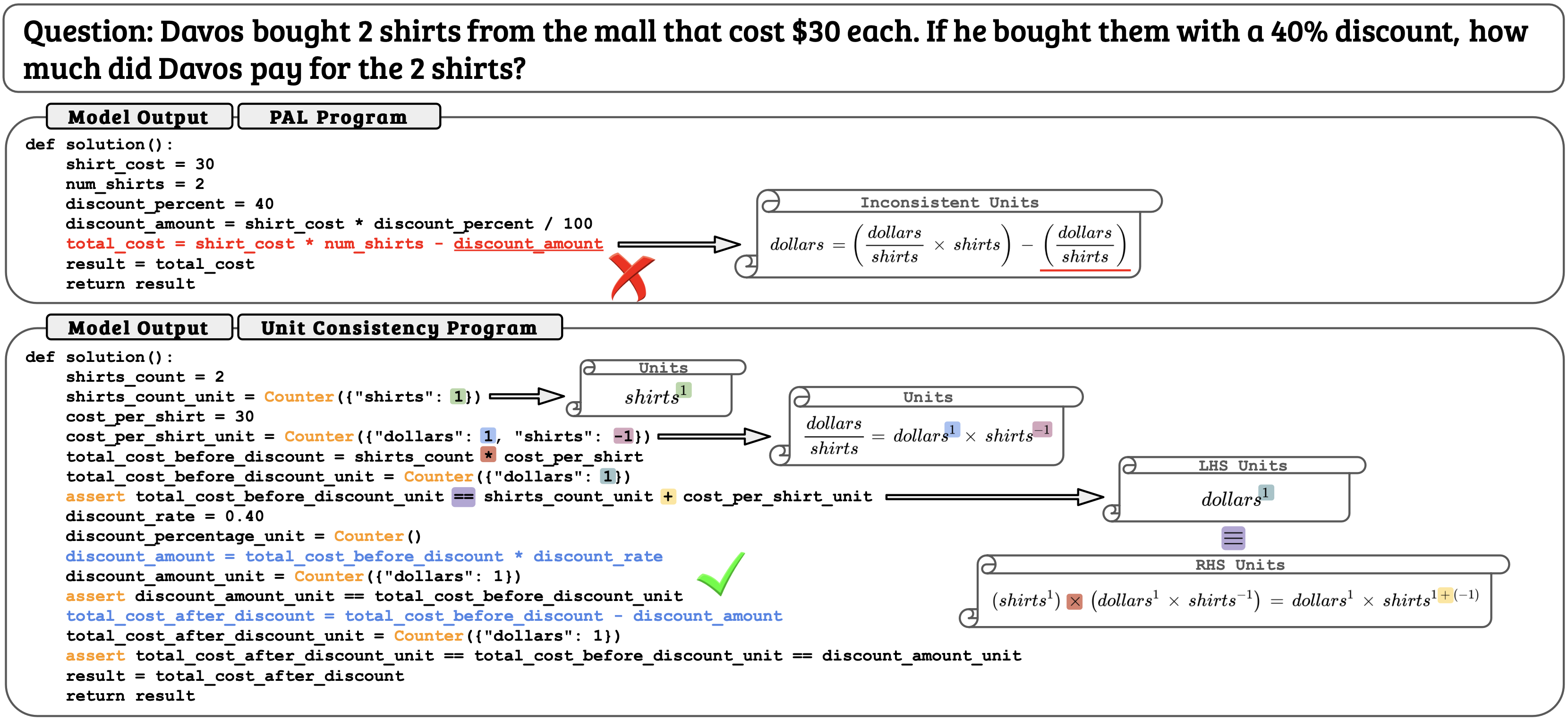}
    \caption{Comparison between PAL-based Programs and Unit Consistency Programs. Unit Consistency Programs contain unit specifications using \counter{} objects and unit verification routines using \assert{} statements.}
    \label{fig:Unit Consistency Program}
\end{figure*}

We propose a methodological framework to enhance the reasoning capabilities of LLMs by introducing a unit system for quantities and enforcing unit consistency. Ensuring unit consistency is crucial for accurate solutions in the context of mathematical word problems. To achieve this, we introduce Unit Consistency Programs (UCPs) (\Cref{fig:Unit Consistency Program}) designed to enhance LLMs’ reasoning abilities by enabling them to self-verify unit consistency within equations. UCPs consist of \counter{} objects responsible for tracking variable units and \assert{} statements generated following each equation involving an operation. These \assert{} statements verify the consistency of units within the equation and can trigger an \assert{} error when inconsistent units are detected. 

We have developed a dataset that pairs math word problems with unit consistency programs containing unit specifications and verification routines. Our preliminary study presents the outcomes of fine-tuning \llama, \code{} , and \mistral{}  using these programs. Although our approach, which incorporates unit consistency, currently slightly underperforms compared to a non-unit-consistent approach, we conducted an in-depth error analysis to understand the reasons behind this discrepancy and proposed several options for future improvements.

\section{Motivating Analysis}

Recent studies have utilized the concept of program-based prompting to generate pseudo-gold programs as an intermediary step for training smaller models \cite{jie2023leveraging, zhu2023pad}. Notably, this approach has shown promising outcomes, although these results still fall short of the performance achieved by larger models like GPT-4 \cite{openai2023gpt4}. To better comprehend the gaps in the mathematical reasoning abilities of smaller models, we fine-tuned \llama{} \cite{touvron2023llama}, \code{} \cite{rozière2023code}, and \mistral{} \cite{jiang2023mistral} using the GSM8K-PAL dataset provided by \citet{jie2023leveraging}, and conducted a comprehensive analysis of the fine-tuned models. The GSM8K-PAL dataset contains approximately 6.8k word problems paired with their PAL annotations in the training dataset as shown in \Cref{tab:dataset uc}.

After fine-tuning these models on GSM8K-PAL, we observed that they struggle with math word problems involving multiple different units. As illustrated in \Cref{fig:Unit Consistency Program} (top), the example illustrates a unit mismatch in the model trained on the PAL-based approach. Specifically, the subtraction operation between variables $\texttt{shirt\_cost} \times \texttt{num\_shirts}$ and $\texttt{discount\_amount}$ is incorrect. The units are incompatible: the former is in \si{\text{dollar}}, and the latter is in \si{\text{dollar}\per\text{shirt}}.

To support our observation that the model struggles with problems containing multiple units, we employed GPT-3.5 Turbo\footnote{GPT-3.5 Turbo annotations were obtained in September 2023.} to categorize the examples from both the train and test splits into two distinct groups. The first group comprises of questions involving a single unit, while the second group comprises of questions with multiple units. This classification was achieved using few-shot prompting, with GPT-3.5 Turbo serving as the backend engine. The specifics of the few-shot prompt utilized are detailed in \Cref{Classification Prompt}, and the distribution of these categories is presented in \Cref{tab:class split}. Our analysis reveals that approximately 40\% of the problems in both training and test splits involve multiple units. 

To further evaluate the accuracy of GPT-3.5 Turbo in identifying questions with multiple units, we conducted a small-scale human assessment, detailed in \Cref{human evaluation}. The first author manually annotated 100 randomly selected test examples from GSM8K and compared the annotations with the classifications made by GPT-3.5 Turbo. The results demonstrated a precision of 80.4\%, indicating that GPT-3.5 Turbo generally excels in predicting questions involving multiple units. We have extended this analysis to the SVAMP \cite{patel-etal-2021-nlp}, as presented in \Cref{SVAMP split}, to demonstrate that this phenomenon is not exclusive to GSM8K.

Based on the test dataset split we collected, we divided the accuracy of the fine-tuned models into two categories: one for questions with a single unit and another for questions with multiple units. This categorization is shown in \Cref{tab:Main Results Table}. A detailed examination of \Cref{tab:Main Results Table} reveals that our observations remained consistent across all three fine-tuned models, indicating superior performance on single-unit problems compared to those with multiple units. Motivated by these findings, we developed Unit Consistency Programs (UCPs) aimed at addressing the limitations inherent in PAL-based solutions.

\begin{table}[t!]
	\centering
    \footnotesize
	\adjustbox{max width=1.0\linewidth}{
		\begin{tabular}{lcccc}
			\toprule
			\multirow{1}{*}{\textbf{Dataset}} &  \multirow{1}{*}{\#\textbf{Train}} &\textbf{\#Program} & \multirow{1}{*}{\#\textbf{Valid}} & \multirow{1}{*}{\#\textbf{Test}}  \\
			\midrule
            GSM8K-PAL & \textcolor{white}{0}7,473 & 6,877 (92.0\%) & - & 1,319 \\ 
			UCPs & \textcolor{white}{0}7,473 & 4,480 (59.9\%) & - & 1,319 \\ 
			\bottomrule
		\end{tabular}
	}
	\caption{Comparison of dataset size of GSM8K-PAL by \cite{jie2023leveraging} and UCPs.}
	\label{tab:dataset uc}
\end{table}

\begin{table}[t]
    \centering
    \footnotesize
        \begin{tabular}{cccc}
            \toprule
            \multicolumn{2}{c}{\textbf{Train Dataset} (7473)} & \multicolumn{2}{c}{\textbf{Test Dataset} (1319)} \\
            \cmidrule(lr){1-2} \cmidrule(lr){3-4}
            \textbf{Single} & \textbf{Multiple} & \textbf{Single} & \textbf{Multiple} \\
            \midrule
            4479 & 2994 & 755 & 564 \\
            (59.9\%) & (40.1\%) & (57.2\%) & (42.8\%) \\
            \bottomrule
        \end{tabular}
    \caption{Classification of GSM8K into two categories: single unit and multiple units.}
    \label{tab:class split}
\end{table}

\section{Methodology}

\subsection{Unit Consistency Programs}
Unit consistency checks are essential safeguards, helping to identify and prevent errors from inconsistent units in mathematical equations. In contrast to PAL/PoT approaches that directly generate programs to solve math word problems, our method enhances these programs by integrating specialized \counter{} objects. These objects are responsible for tracking variable units and ensuring the correct handling of operations with differing units. Additionally, we incorporate \assert{} statements after each equation, as illustrated in \Cref{fig:Unit Consistency Program} (bottom). These \assert{} statements verify unit consistency within equations, triggering an error if unit mismatches are detected.

Consider the example in \Cref{fig:Unit Consistency Program} (bottom), illustrating a multiplication operation between \texttt{shirts\_count} (measured in `shirts') and \texttt{cost\_per\_shirt} (measured in `dollars per shirt'). In this operation, the units of `shirts' from \texttt{shirts\_count} and `per shirt' from \texttt{cost\_per\_shirt} naturally cancel each other out, resulting in a unit of `dollars'. An \assert{} statement is used to verify this expected cancellation of units. In our notation, the exponent of a unit in the numerator is represented as +1, and in the denominator as -1. Therefore, in this multiplication, the positive exponent of `shirts' in \texttt{shirts\_count} cancels with the negative exponent of `per shirt' in \texttt{cost\_per\_shirt}, aligning the product's right-hand side (RHS) with the expected left-hand side (LHS) unit of \texttt{total\_cost\_before\_discount}, confirming it is in `dollars'. The example also illustrates a unitless quantity, specifically a percentage. In this case, there won't be any units specified in the \counter{} initialization. Our methodology requires the development of a specialized \counter{} class, details of which are elaborated in the \Cref{collections Counter}.

\begin{table}[t]
  \centering
  \footnotesize
  \resizebox{0.48\textwidth}{!}{
  \begin{tabular}{lccc}
    \toprule \rowcolor{white}
     & \textbf{Positive Predicted} & \textbf{Negative Predicted} \\
    \midrule
    \textbf{Actual Positive} & 37 & 16\\
    \textbf{Actual Negative} & 9 & 38\\
    \toprule
    \multicolumn{1}{c}{\textbf{Precision}} & \textbf{Recall} & \textbf{Accuracy} \\
    \multicolumn{1}{c}{80.4\%} & 69.8\% & 75.0\% \\
    \bottomrule
  \end{tabular}
  }
  \caption{Small human evaluation compared on GPT-3.5 Turbo classification on 100 randomly sampled test examples from GSM8K. Human annotations were done by the first author.}
  \label{human evaluation}
\end{table}

\subsection{Training Data Annotations}
Adopting the methodology used in PAL/PoT, we sampled programs for each math word problem, adding them to our training data when their execution yielded the correct answer. For each math word problem $x$ in the training dataset $D$, we performed greedy decoding at temperature $T = 0$ to synthesize program $P_{ucp}$. Upon executing the program $P_{ucp}$, if the predicted answer $\hat{y}$ matched the ground-truth answer $y$ and $P_{ucp}$ consists of \counter{} objects and \assert{} statements, we included the tuple $(x, P_{ucp}, y)$ in our new training dataset $D_{ucp}$. Any math word problem $x$ for which a matching program couldn't be obtained was discarded.

\subsection{Fine-tuning Small Models}
We fine-tuned smaller models with our annotated dataset $D_{ucp}$ through standard causal language modeling techniques. The objective is to generate a corresponding Python program $P$ for a given math word problem $x$. After fine-tuning, the model was used to generate Python programs, which were then executed using a Python interpreter to obtain the final answer. We employed strong open-source LLMs such as \llama, \code, and \mistral{} as our models to fine-tune.

\section{Experiments}

\subsection{Dataset}
We conducted our experiments primarily on GSM8K, employing few-shot prompting with GPT-4 for the first 1,000 examples\footnote{GPT-4 annotations obtained in September 2023.} and GPT-4 Turbo for the remaining 6,473 examples\footnote{GPT-4 Turbo annotations obtained in December 2023.} in the GSM8K train dataset. We used six manually crafted Unit Consistency Programs (UCPs) samples, as detailed in \Cref{Program Synthesis Prompt}. We successfully annotated 59.9\% of the GSM8K train dataset, creating our annotated UCPs dataset, $D_{ucp}$. \Cref{tab:dataset uc} presents the statistics of our UCPs dataset.

\subsection{Baseline}
Our baseline models consist of different models such as \llama, \code, and \mistral{} fine-tuned on GSM8K-PAL. We use this as a direct baseline to our method as it provides a more effective comparison between our method UCPs and existing methods like PAL/POT since our UCPs serve as extensions to typical Python programs used for solving mathematical problems, as demonstrated in PAL/POT.

\begin{table}[t]
    \centering
    \footnotesize
    \adjustbox{max width=1.0\linewidth}{
    \begin{tabular}{l|cc|c}
        \toprule
        \textbf{Model} & \textbf{Single} & \textbf{Multiple} & \textbf{Overall} \\
        \midrule
         \multicolumn{4}{c}{\textit{Closed-Source Models}} \\
         \midrule
        GPT-4  & - & - & 92.0 \\
        GPT-3.5-Turbo & - & - & 80.8 \\
        \midrule
         \multicolumn{4}{c}{\textit{Open-Source Models 7B}} \\
         \midrule

        Llama-2 (PAL)$^\dag$  & 58.5 \ua{3.1} & 51.2 \da{4.2} & 55.4 \\
        Code-Llama (PAL)$^\dag$  & 65.6 \ua{2.5} & 59.8 \da{3.3} & 63.1 \\
        Mistral (PAL)$^\dag$  & 72.2 \ua{1.8} & 68.1 \da{2.3} & 70.4 \\ \rowcolor{gray}
        \hline
        VerityMath-Llama-2 & 51.9 \ua{5.7} & 38.7 \da{7.5} & 46.2 \\ \rowcolor{gray}
        VerityMath-Code-Llama & 58.4 \ua{4.2} & 48.6 \da{5.6} & 54.2 \\ \rowcolor{gray}
        VerityMath-Mistral & 71.5 \ua{3.3} & 63.7 \da{4.5} & 68.2 \\ 
        \bottomrule
    \end{tabular}
    }
    \caption{Comparison of test accuracy on GSM8K of different 7B open-source models fine-tuned on PAL and UCP. The \colorbox{green!10}{green} and \colorbox{red!10}{red} boxes represent the increase and decrease in accuracy compared to its overall score. $^\dag$We fine-tune the model using GSM8K-PAL by \citet{jie2023leveraging}.} 
    \label{tab:Main Results Table}
\end{table}

\subsection{Implementation}
We conducted fine-tuning experiments on GSM8K-PAL and UCPs, details of both datasets can be found in \Cref{tab:dataset uc}. In our fine-tuning experiments, we utilized the QLoRA technique \cite{dettmers2023qlora} for enabling efficient fine-tuning. All QLoRA hyper-parameters were set as presented in \citet{dettmers2023qlora}. In all our experiments we use NF4 with double quantization and bf16 computation datatype. We set LoRA $r\!=\!64,\, \alpha\!=\!16$, and add LoRA modules on all linear layers of the base model. We also use max grad norm of 0.3 and LoRA dropout of 0.1. We use AdamW optimizer and set the learning rate to $2e^{-4}$, with a batch size of $32$ and a maximum context length of $1024$. We trained the model for $20$ epochs using $4$ A100 40 GB GPUs which took roughly 14 hours and evaluated it on the test dataset.

\subsection{Main Results}
Our model, VerityMath-\mistral, fine-tuned on UCPs achieved an overall accuracy of $68.2\%$ on the GSM8K test dataset. Specifically, it attained $71.5\%$ accuracy for problems involving a \emph{single} unit and $63.7\%$ accuracy for those with \emph{multiple} units, as detailed in \Cref{tab:Main Results Table}. 
When compared to the \mistral{} (PAL) baseline, VerityMath-\mistral{} exhibits a slight overall accuracy decrease of 2.2\%. Meanwhile, VerityMath-\code{} and VerityMath-\llama{} experienced more significant declines in their overall accuracy, approximately 9\% lower than their respective PAL counterparts.
Specifically, VerityMath-Code-Llama achieved 54.2\% overall accuracy, with 58.4\% for single unit problems and 48.6\% for multiple units, while VerityMath-Llama-2 achieved an overall accuracy of 46.2\%, with 51.9\% for single unit and 38.7\% for multiple units.

\begin{figure}[t]
 \centering
 \includegraphics[width=0.48\textwidth]{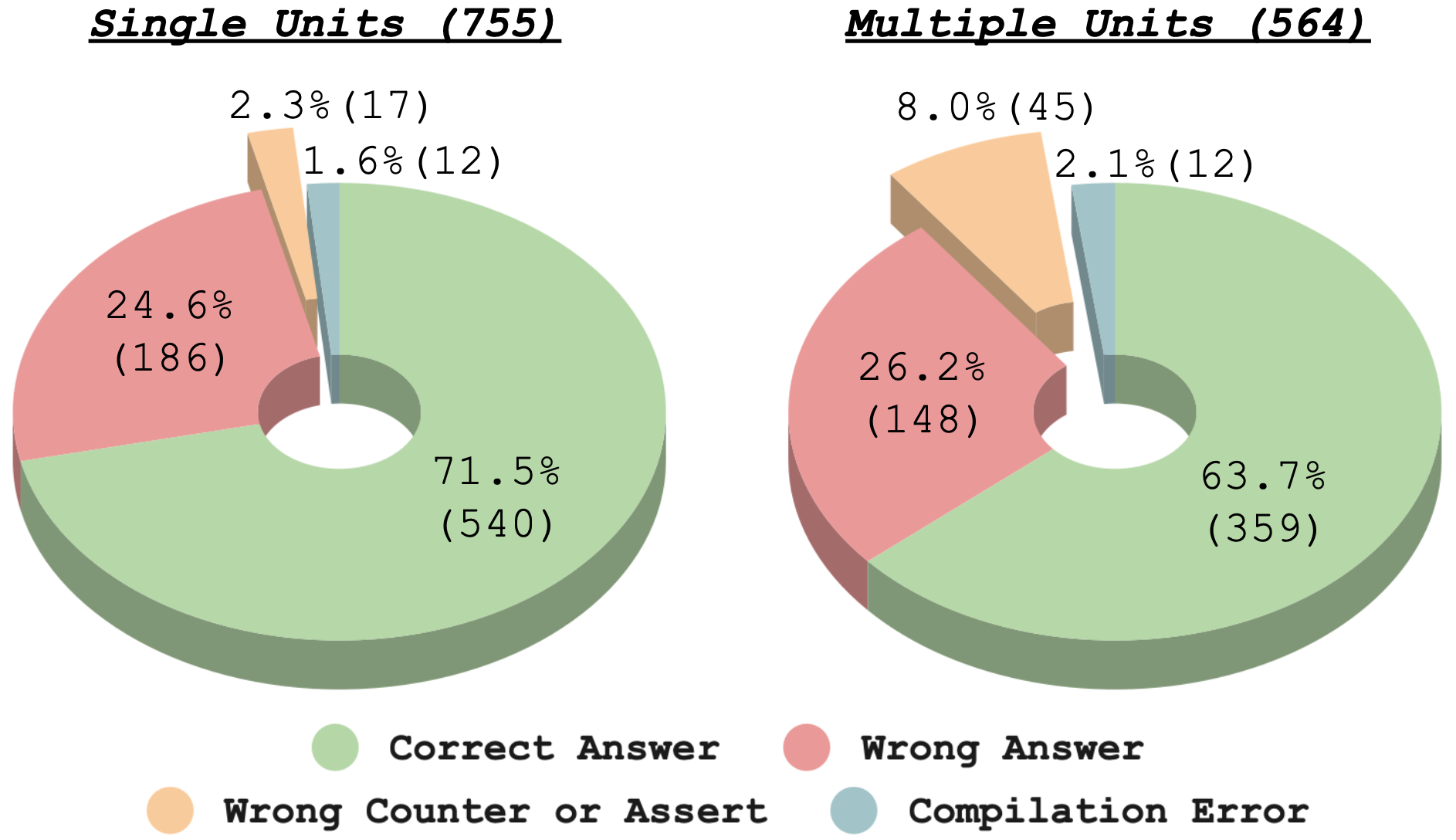}
     \caption{Error analysis of VerityMath-\mistral. Correct Answer: The program compiles and produces the correct answer. Wrong Answer: The program compiles but produces an incorrect answer. Wrong \counter{} or \assert{} : After removing \counter{} and \assert{} statements, the program produces the correct answer. Compilation Error: The program is unable to compile.}
     \label{fig:error analysis}
 \end{figure}

\subsection{Analysis}
In this section, we conducted an in-depth analysis of the potential causes for the decline in overall accuracy in the GSM8K test dataset. We focused on VerityMath-\mistral{} for all of our analysis.

\paragraph{Correctness of \counter{} and \assert{} statements}
In an error analysis of VerityMath-\mistral{} outputs from the test dataset, we observed some challenges that led to decreased performance, specifically, the correctness of \counter{} and \assert{} statements. 
We reran the whole evaluation but this time, when we were met with a program that raised an assertion error, we removed the \counter{} and \assert{} statements and executed the programs again.
If the program compiles and produces the correct answer after this modification, it indicates that the program was originally incorrect due to incorrect \counter{} or \assert{} statements.
Referring to \Cref{fig:error analysis}, we observed a notable percentage of output programs that contained incorrect \counter{} or \assert{} statements in VerityMath-\mistral{} outputs. Specifically, $2.3\%$ of the problems with single units and $8.0\%$ of the problems with multiple units have incorrect \counter{} and \assert{} which caused correct programs that would have resulted in the correct answer to have a false assertion error resulting in the wrong answer. 
Examples of such cases with incorrect \counter{} and \assert{} are shown in \Cref{incorrect UCPs}.

\begin{figure}[t]
 \centering
 \includegraphics[width=0.48\textwidth]{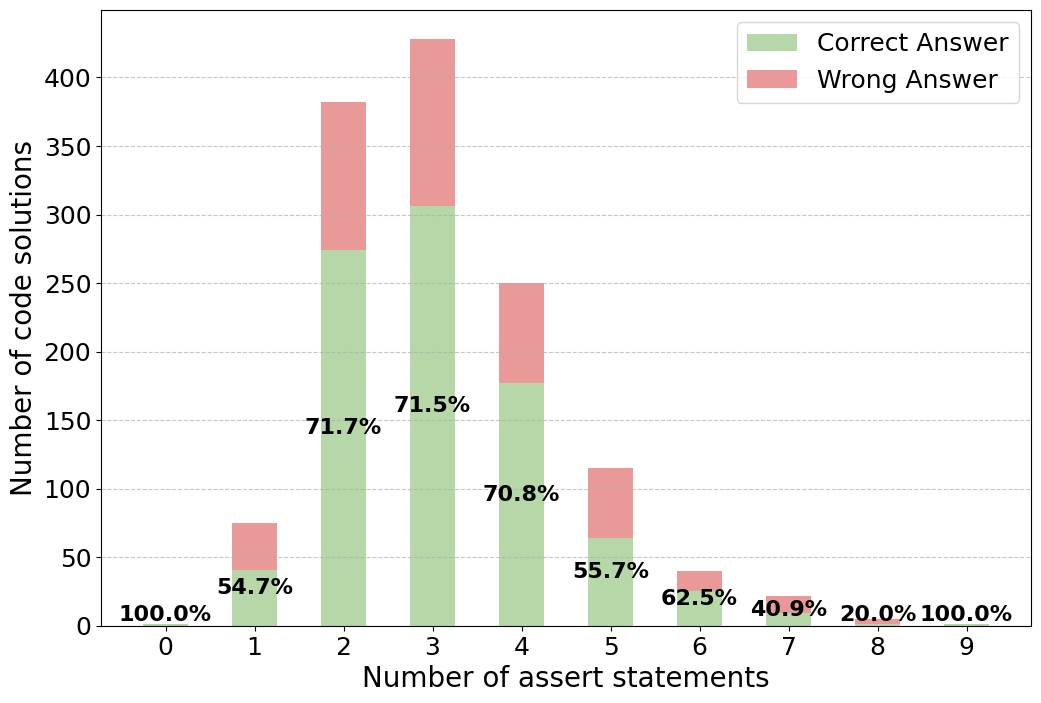}
     \caption{Performance of VerityMath-\mistral{} on the GSM8K test dataset based on the number of \assert{} statements in the code solution. The percentage shown in each bar represents the percentage of correct answers given the number of \assert{} statements in the code solution.}
     \label{fig:number assert}
 \end{figure}

\paragraph{Performance based on number of \assert{} statements}
We further conducted a detailed analysis of code solutions categorized by the number of assert statements, as shown in \Cref{fig:number assert}. 
Each bar represents the total number of code solution that consists of a specific number of assert statements. 
The green segments of the bars indicate the count of code solutions that resulted in the correct answer, while the red segments represent those that resulted in an incorrect answer. The percentage of correct answers is annotated on each bar for clarity.
It is evident from the plot that the percentage of correct answers generally decreases as the number of assert statements increases, from code solutions with 2 to 4 assert statements having approximately 70\% accuracy to code solutions with 5, 6, and 7 assert statements having 55.7\%, 62.5\%, and 40.9\% respectively.
Highlighting a trend where more complex code solutions with more assert statements are more likely to result in incorrect answers.
This aligns with the earlier observations regarding the correctness of \assert{} statements, and suggests that with more assert statements in the code solution, it is more prone to having errors due to the incorrect \assert{} statements which would then result in a wrong answer.

\begin{figure}[t]
 \centering
 \includegraphics[width=0.48\textwidth]{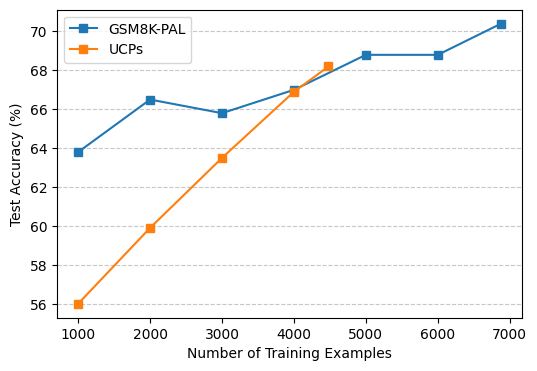}
     \caption{Performance of VerityMath-\mistral{} as we scale the number of training examples of GSM8K-PAL and UCPs. GSM8K-PAL has a total of 6877 annotated training examples whereas UCPs have 4480 annotated training examples.}
     \label{fig:training examples}
 \end{figure}

\paragraph{Impact of training annotations}
Due to the difference in the number of training examples between GSM8k-PAL and UCPs of 2397 as shown in \Cref{tab:dataset uc}. It is crucial to also understand the implications of the number of training examples with respect to the performance. We fine-tuned \mistral{} on both GSM8k-PAL and UCPs with an interval of 1000 training examples and showed the results in \Cref{fig:training examples}. The performance of \mistral{}  when fine-tuned on GSM8k-PAL or UCPs demonstrates a clear trend of improvement with the increase in the number of training examples.
For GSM8k-PAL, the test accuracy starts at 63.8\% with 1,000 training examples and steadily increases to 70.4\% with 6,877 examples. On the other hand, The UCPs exhibit a more pronounced improvement curve, starting at 56.0\% accuracy with 1,000 training examples, the performance increases significantly to 68.2\% with 4,480 examples. This rate of improvement indicates that with limited examples, the concept of UCPs is harder to grasp for \mistral{} as compared to PAL.
The difference in performance gains suggests that UCPs might have untapped potential that could be realized with an increased number of training examples and it implies that with sufficient training examples, UCPs could potentially surpass PAL in performance.

In our in-depth anaysis, we identified a notable bottleneck in our current method, which is the correctness of \counter{} and \assert{} statements. This issue led to a slight decrease in performance. Our method, UCPs, is a relatively more complex method for existing 7B LLMs to learn, but with a significant increase in dataset annotations, it is highly possible that our method will outshine the existing PAL method. 
Another approach could involve data augmentation using synthetic examples \cite{pmlr-v139-wu21c}.
Examples showcasing the efficacy of UCPs are available in \Cref{correct UCPs}.

\section{Related Work}

Our research builds upon the Program of Thoughts (PoT) approach \cite{chen2022program} and the Program Aided Language Model (PAL) \cite{gao2023pal} methodologies, which have shown effectiveness in solving mathematical problems. These approaches have outperformed techniques like the Chain-of-Thought (CoT) \cite{wei2023chainofthought}, which can struggle with computational inaccuracies \cite{lewkowycz2022solving}. We extend their work by focusing on the use of programs for solving math word problems and the concept of self-verification to improve LLMs' reasoning capabilities.

The advancement of GPT models \cite{brown2020language} has inspired various studies \cite{ho2023large, fu2023specializing, magister2023teaching, shridhar2023distilling} on creating synthetic datasets for fine-tuning smaller models \cite{hinton2015distilling}. Notably, \citet{zhu2023pad} used PAL annotations in this context, \cite{magister2023teaching, ho2023large, yu2023metamath} employed CoT annotations, and \citet{yue2023mammoth} used a hybrid of CoT and PoT rationales.

In mathematical problem-solving, ensuring solution validity is crucial due to hallucinations in LLMs \cite{bubeck2023sparks} and challenges in executing multiplications \cite{dziri2023faith}. Prior research has focused on training additional verifiers for answer accuracy \cite{cobbe2021training}, providing feedback for each intermediate reasoning step \cite{lightman2023lets}, and integrating tools to agents\cite{gou2024tora}. However, \citet{weng2023large} and \citet{miao2023selfcheck} have shown potential for LLMs to self-verify solutions. Our approach builds on these insights, incorporating programs for solving math word problems and leveraging self-verification to enhance LLM reasoning.

\section{Conclusion and Future Work}
In this study, we analyzed open-source Large Language Models (LLMs) and pinpointed their struggle with math problems involving multiple units, highlighting a key improvement area. We introduced Unit Consistency Programs (UCPs) as a novel method to address LLMs' reasoning and verification abilities, especially in complex math problems. We identified some limitations in our current approach. Future work will focus on advancing unit check methodologies in UCPs to address these limitations.

\section*{Limitations}
Recent creations of synthetic datasets for math problem-solving often rely on prompting large language models (LLMs), such as GPT-4. However, this approach can be costly, and on a large scale, the expenses escalate. Our dataset creation incurred a total cost of approximately \$350 USD. Due to budget constraints, we couldn't sample multiple reasoning paths per question, as presented in \citet{wang2023selfconsistency}, limiting the potential for increased annotations.

\section*{Impact Statement}
This paper presents work whose goal is to advance the field of math problem-solving using LLMs. However, it is also crucial to be aware of the potential risks associated with VerityMath. Due to the current challenges of VerityMath, the units initialized by \counter{} and \assert{} statements may not always be accurate. Consequently, it is strongly recommended to exercise caution when relying on VerityMath outputs for any use.

\section*{Acknowledgements}
This research is supported by the Ministry of Education, Singapore, under its Science of Learning Grant (award ID: MOESOL2021-0006). Any opinions, findings and conclusions or recommendations expressed in this material are those of the author(s) and do not reflect the views of the Ministry of Education, Singapore. The computational work for this article was partially performed on resources of the National Supercomputing Centre (NSCC), Singapore (\url{https://www.nscc.sg}).


\bibliography{custom}
\bibliographystyle{icml2024}

\newpage
\appendix
\onecolumn
\section{Appendix}
\label{sec:appendix}

\subsection{Manually constructed context samples for program synthesis} \label{Program Synthesis Prompt}

\subsubsection{6-shot Prompt} 

\begin{lstlisting}[basicstyle=\scriptsize, escapechar={|}]
Question: Ashley bought a big bag of 96 candies. Yesterday, she ate 15 candies and today, she ate twice as much candies as yesterday. How many candies were left?

Python solution:
def solution():
    """Ashley bought a big bag of 96 candies. Yesterday, she ate 15 candies and today, she ate twice as much candies as yesterday. How many candies were left?"""
    candies_initial = 96
    candies_initial_unit = Counter({"candies": 1})
    candies_ate_yesterday = 15
    candies_ate_yesterday_unit = Counter({"candies": 1})
    candies_ate_today = candies_ate_yesterday * 2 
    candies_ate_today_unit = Counter({"candies": 1})
    assert candies_ate_today_unit == candies_ate_yesterday_unit
    total_candies_eaten = candies_ate_yesterday + candies_ate_today
    total_candies_eaten_unit = Counter({"candies": 1})
    remaining_candies = candies_initial - total_candies_eaten
    remaining_candies_unit = Counter({"candies": 1})
    assert remaining_candies_unit == candies_initial_unit == total_candies_eaten_unit
    result = remaining_candies
    return result

Question: There are 235 books in the library. On Monday, 122 books were taken out. On Tuesday, half of the books taken on Monday were brought back. How many books are there now?

Python solution:
def solution():
    """There are 235 books in the library. On Monday, 122 books were taken out. On Tuesday, half of the books taken on Monday were brought back. How many books are there now?"""
    books_initial = 235
    books_initial_unit = Counter({"books": 1})
    books_taken_monday = 122
    books_taken_monday_unit = Counter({"books": 1})
    books_remaining_after_monday = books_initial - books_taken_monday
    books_remaining_after_monday_unit = Counter({"books": 1})
    assert books_remaining_after_monday_unit == books_initial_unit == books_taken_monday_unit
    books_brought_back_tuesday = books_taken_monday / 2  
    books_brought_back_tuesday_unit = Counter({"books": 1})
    assert books_brought_back_tuesday_unit == books_taken_monday_unit
    books_remaining_after_tuesday = books_remaining_after_monday + books_brought_back_tuesday
    books_remaining_after_tuesday_unit = Counter({"books": 1})
    assert books_remaining_after_tuesday_unit == books_remaining_after_monday_unit == books_brought_back_tuesday_unit
    result = books_remaining_after_tuesday
    return result

Question: There is a group of 10 people who are ordering pizza. If each person gets 2 slices and each pizza has 4 slices, how many pizzas should they order?

def solution():
    """There is a group of 10 people who are ordering pizza. If each person gets 2 slices and each pizza has 4 slices, how many pizzas should they order?"""
    people_total = 10
    people_total_unit = Counter({"people": 1})
    pizza_slices_per_person = 2
    pizza_slices_per_person_unit = Counter({"slices": 1, "people": -1})
    pizza_slices_total = people_total * pizza_slices_per_person
    pizza_slices_total_unit = Counter({"slices": 1})
    assert pizza_slices_total_unit == people_total_unit + pizza_slices_per_person_unit
    slices_per_pizza = 4
    slices_per_pizza_unit = Counter({"slices": 1, "pizza": -1})
    pizza_total = pizza_slices_total / slices_per_pizza
    pizza_total_unit = Counter({"pizza": 1})
    assert pizza_total_unit == pizza_slices_total_unit - slices_per_pizza_unit
    result = pizza_total
    return result
    
Question: Lana has 2 bags with 2 marbles in each bag. Markus has 2 bags with 3 marbles in each bag. How many more marbles does Markus have?

Python solution:
def solution():
    """Lana has 2 bags with 2 marbles in each bag. Markus has 2 bags with 3 marbles in each bag. How many more marbles does Markus have?"""
    bags_lana = 2
    bags_lana_unit = Counter({"bags": 1})
    marbles_per_bag_lana = 2
    marbles_per_bag_lana_unit = Counter({"marbles": 1, "bags": -1})
    marbles_total_lana = bags_lana * marbles_per_bag_lana
    marbles_total_lana_unit = Counter({"marbles": 1})
    assert marbles_total_lana_unit == marbles_per_bag_lana_unit + bags_lana_unit
    bags_markus = 2
    bags_markus_unit = Counter({"bags": 1})
    marbles_per_bag_markus = 3
    marbles_per_bag_markus_unit = Counter({"marbles": 1, "bags": -1})
    marbles_total_markus = bags_markus * marbles_per_bag_markus
    marbles_total_markus_unit = Counter({"marbles": 1})
    assert marbles_total_markus_unit == marbles_per_bag_markus_unit + bags_markus_unit
    marbles_more_markus = marbles_total_markus - marbles_total_lana
    marbles_more_markus_unit = Counter({"marbles": 1})
    assert marbles_more_markus_unit == marbles_more_markus_unit == marbles_total_lana_unit
    result = marbles_more_markus
    return result
    
Question: Sally has 4 containers with the same amount of cookies in them, totaling 12 cookies. John has 4 containers with the same amount of cookies in them, totaling 24 cookies. How many more cookies does John have in each container?

Python solution:
def solution():
    """Sally has 4 containers with the same amount of cookies in them, totaling 12 cookies. John has 4 containers with the same amount of cookies in them, totaling 24 cookies. How many more cookies does John have in each container?"""
    containers_sally = 4
    containers_sally_unit = Counter({"containers": 1})
    total_cookies_sally = 12
    total_cookies_sally_unit = Counter({"cookies": 1})
    cookies_per_container_sally = total_cookies_sally / containers_sally
    cookies_per_container_sally_unit = Counter({"cookies": 1, "containers": -1})
    assert cookies_per_container_sally_unit == total_cookies_sally_unit - containers_sally_unit
    containers_john = 4
    containers_john_unit = Counter({"containers": 1})
    total_cookies_john = 24
    total_cookies_john_unit = Counter({"cookies": 1})
    cookies_per_container_john = total_cookies_john / containers_john
    cookies_per_container_john_unit = Counter({"cookies": 1, "containers": -1})
    assert cookies_per_container_john_unit == total_cookies_john_unit - containers_john_unit
    more_cookies_per_container_john = cookies_per_container_john - cookies_per_container_sally
    more_cookies_per_container_john_unit = Counter({"cookies": 1, "containers": -1})
    assert more_cookies_per_container_john_unit == cookies_per_container_john_unit == cookies_per_container_sally_unit
    result = more_cookies_per_container_john
    return result

Question: It takes Peter 4 hours to fix his fence. He spends 20 minutes everyday fixing the fence, how many days will it take for Peter to finish fixing the fence?

Python solution:
def solution():
    """It takes Peter 4 hours to fix his fence. He spends 20 minutes everyday fixing the fence, how many days will it take for Peter to finish fixing the fence?"""
    hours_to_fix_fence = 4
    hours_to_fix_fence_unit = Counter({"hours": 1})
    minutes_per_hour = 60
    minutes_per_hour_unit = Counter({"minutes": 1, "hours": -1})
    minutes_to_fix_fence = hours_to_fix_fence * minutes_per_hour
    minutes_to_fix_fence_unit =  Counter({"minutes": 1})  
    assert minutes_to_fix_fence_unit == hours_to_fix_fence_unit + minutes_per_hour_unit
    minutes_per_day_to_fix_fence = 20
    minutes_per_day_to_fix_fence_unit = Counter({"minutes": 1, "days": -1})
    total_days_to_fix_fence = minutes_to_fix_fence / minutes_per_day_to_fix_fence
    total_days_to_fix_fence_unit = Counter({"days": 1})
    assert total_days_to_fix_fence_unit == minutes_to_fix_fence_unit - minutes_per_day_to_fix_fence_unit
    result = total_days_to_fix_fence
    return result
\end{lstlisting}

\subsubsection{System Prompt} 

You are an experienced professional skilled in using python programs to solve math problems and enhancing code robustness through the strategic implementation of unit counters and assert statements to ensure flawless unit-level consistency checks. You will only write code blocks.

\subsection{Manually constructed context samples for Classification (Single Unit/ Multiple Units)} \label{Classification Prompt}

\subsubsection{8-shot Prompt} 

\noindent\textbf{Question}: In ten years, Samantha will be three times as old as Alex is today. If Alex is currently 14 years old, how old is Samantha now? 

\noindent\textbf{Explanation}:  

\noindent In this problem, a unit consistency check is not required to ensure the accuracy of measurements and calculations. We are dealing with a single unit - age (years). The calculations involve simple addition, subtraction, and multiplication of the same unit, so there is no need for unit conversions or comparisons between different units. In this case, all the quantities are given in terms of years, and there are no conflicting units or measurements that would require a unit consistency check. Answer: No. \\

\noindent\textbf{Question}: Sarah has a collection of stickers. Every alternate day, she adds 3 new stickers to her collection for 4 weeks. For each sticker, she spends 5 minutes sticking it onto her collection book. How many minutes did she spend sticking stickers? 

\noindent\textbf{Explanation}: 

\noindent In this problem, a unit consistency check is required to ensure the accuracy of measurements and calculations. We are dealing with different units - the number of days, the number of stickers, and the time spent sticking each sticker (minutes). By performing a unit consistency check, we can ensure that the conversions between different units are accurate and that the final result makes sense in the context of the problem. In this case, we can calculate the total number of stickers added by multiplying the number of stickers added per day by the number of days, and then calculate the total time spent sticking stickers by multiplying the number of stickers by the time spent sticking each sticker. Answer: Yes. \\

\noindent\textbf{Question}: Suzie practices piano every weekday for 3 weeks. Each day, she practices 2 songs that are 4 minutes long and 1 song that is 3 times longer than the other two songs combined. How many minutes does Suzie practice in total during these 3 weeks? 

\noindent\textbf{Explanation}: 

\noindent In this problem, a unit consistency check is required to ensure the accuracy of measurements and calculations. We are dealing with different units - the number of days, the number of songs, and the time spent practicing each song (minutes). By performing a unit consistency check, we can ensure that the conversions between different units are accurate and that the final result makes sense in the context of the problem. In this case, we can calculate the total time spent practicing each day by adding the time spent practicing the two shorter songs and the longer song, and then calculate the total time spent practicing over the three weeks by multiplying the daily practice time by the number of days. Answer: Yes. \\

\noindent\textbf{Question}: A girl rabbit hops along a garden path and picks 8 carrots. She places them in a basket with 16 other carrots she has already collected. While on her way home, the girl rabbit accidentally drops 2 carrots. How many carrots are left in the basket? 

\noindent\textbf{Explanation}:  

\noindent In this problem, a unit consistency check is not required to ensure the accuracy of measurements and calculations. We are dealing with a single unit - the number of carrots. The calculations involve simple addition and subtraction of the same unit, so there is no need for unit conversions or comparisons between different units. In this case, all the quantities are given in terms of the number of carrots, and there are no conflicting units or measurements that would require a unit consistency check. Answer: No. \\

\noindent\textbf{Question}: A bottle of shampoo is 500ml and each hair wash uses 20 ml of shampoo. Each salon charges \$3.00 for one hair wash. How much money does a salon make on a bottle of shampoo? 

\noindent\textbf{Explanation}:  

\noindent In this problem, a unit consistency check is required to ensure the accuracy of measurements and calculations. We are dealing with different units - the volume of shampoo (milliliters), the volume of shampoo used per wash (milliliters), and the cost of each wash (dollars). By performing a unit consistency check, we can ensure that the conversions between different units are accurate and that the final result makes sense in the context of the problem. In this case, we can calculate the total number of hair washes per bottle by dividing the total volume of the bottle by the volume used per hair wash, and then calculate the total money made by multiplying the number of hair washes by the cost of each hair wash. Answer: Yes. \\

\noindent\textbf{Question}: Natalie and Liam have a total of 72 stickers. However, Natalie has half the number of stickers as compared to Liam. How many stickers does Liam have? 

\noindent\textbf{Explanation}:  

\noindent In this problem, a unit consistency check is not required to ensure the accuracy of measurements and calculations. We are dealing with a single unit - the number of stickers. The calculations involve simple addition, subtraction, multiplication, and division of the same unit, so there is no need for unit conversions or comparisons between different units. In this case, all the quantities are given in terms of the number of stickers, and there are no conflicting units or measurements that would require a unit consistency check. Answer: No. \\

\noindent\textbf{Question}: Emma makes \$35,000 a year. She spends \$12,000 on rent every year, \$3,000 on groceries every year, and \$4,500 on gas every year. If she wants to have at least \$15,000 left, how much more money must she make each year? 

\noindent\textbf{Explanation}:  

\noindent In this problem, a unit consistency check is not required to ensure the accuracy of measurements and calculations. We are dealing with a single unit - money (dollars). The calculations involve simple addition and subtraction of the same unit, so there is no need for unit conversions or comparisons between different units. In this case, all the quantities are given in terms of dollars, and there are no conflicting units or measurements that would require a unit consistency check. Answer: No. \\

\noindent\textbf{Question}: Maria pays \$35/hour for 4 hours per week of flute lessons and \$22/hour for 6 hours a week of guitar lessons. How much more does she spend on guitar lessons than flute lessons in a year? 

\noindent\textbf{Explanation}: 

\noindent In this problem, a unit consistency check is required to ensure the accuracy of measurements and calculations. We are dealing with different units - the cost per hour (dollars/hour) and the number of hours per week. By performing a unit consistency check, we can ensure that the conversions between different units are accurate and that the final result makes sense in the context of the problem. In this case, we can calculate the total cost of flute lessons per week by multiplying the cost per hour by the number of hours, and then calculate the total cost of guitar lessons per week by multiplying the cost per hour by the number of hours. Finally, we can calculate the difference in cost between guitar and flute lessons in a year by subtracting the total cost of flute lessons from the total cost of guitar lessons and multiplying it by the number of weeks in a year. Answer: Yes.

\subsubsection{System Prompt} 
Create a coherent explanation for the importance of performing unit consistency checks in various mathematical problems and identify whether a question requires verification of unit consistency. Be as detailed as possible and write your explanation in 1 paragraph and end with Answer: Yes / No.

\subsection{Breakdown of SVAMP dataset} \label{SVAMP split}

\begin{table}[H]
    \centering
    \footnotesize
        \begin{tabular}{cccc}
            \toprule
            \multicolumn{2}{c}{\textbf{Train Dataset} (192)} & \multicolumn{2}{c}{\textbf{Test Dataset} (81)} \\
            \cmidrule(lr){1-2} \cmidrule(lr){3-4}
            \textbf{Single} & \textbf{Multiple} & \textbf{Single} & \textbf{Multiple} \\
            \midrule
            102 & 90 & 34 & 47 \\
            (53.1\%) & (46.9\%) & (42.0\%) & (58.0\%) \\
            \bottomrule
        \end{tabular}
    \caption{SVAMP Dataset split. We only considered the portion which has type Multiplication or Common-Division.}
    \label{tab:SVAMP class split}
\end{table}

The SVAMP dataset comprises a total of 1000 examples, with 700 allocated to the train dataset and 300 to the test dataset. The dataset encompasses four problem types: subtraction, addition, common-division, and multiplication. However, our analysis focuses solely on multiplication and common-division, as problems involving only addition or subtraction are defined to only consist of a single unit. We can observe from \ref{tab:SVAMP class split} that 46.9\% and 58\% of the problems are classified as multiple units in the train and test dataset respectively.

\subsection{Details on Unit Consistency Programs}

\subsubsection{Usage of \counter{} Class} \label{usage Counter}

\begin{lstlisting}[basicstyle=\scriptsize]
Counter({"marbles": 1, "bags":-1}  # marbles per bag

Counter({"slices": 1, "pizza":-1}  # slices per pizza

Counter({"books"}: 1)  # books

Counter()  # percentage (unitless)


\end{lstlisting}

\subsubsection{Undesirable behavior of collections Counter} \label{collections Counter}
Our work necessitated the creation of a custom \counter{} class. In programming, a `class' is a blueprint for creating objects with specific attributes and behaviors. The standard \counter{} class, part of Python's \texttt{collections} module-a collection of specialized data types-was not fully suitable for our needs. It had the limitation of only counting positive values, meaning it would not accurately reflect counts when dealing with negative or zero values. Therefore, we had to create our own custom \counter{} class to handle this limitation.

\begin{lstlisting}[language=Python, basicstyle=\scriptsize, escapechar={|}]
from collections import Counter

Counter({"miles": 1}) - Counter({"hour": 1})
>>> Counter(|\texttt{\color{red}{\{'miles': 1\}}}|)

Desired Output >>> Counter(|\texttt{\color{green}{\{'miles': 1, "hour": -1\}}}|)

\end{lstlisting}

\subsubsection{Assert Statement Formulation} \label{assert formulation}

\begin{lstlisting}[basicstyle=\scriptsize]
Formulation of assert statement for equations with multiplication:
pizza_slices_total = people_total * pizza_slices_per_person
assert pizza_slices_total_unit == people_total_unit + pizza_slices_per_person_unit

Formulation of assert statement for equations with division:
total_cups_sugar = total_teaspoons_sugar / teaspoons_per_cup_sugar
assert total_cups_sugar_unit == total_teaspoons_sugar_unit - teaspoons_per_cup_sugar_unit

Formulation of assert statement for equations with addition:
total_money = allowance + extra_money
assert total_money_unit == allowance_unit == extra_money_unit

Formulation of assert statement for equations with subtraction:
apps_remaining_after_deletion = apps_initial - apps_deleted
assert apps_remaining_after_deletion_unit == apps_initial_unit == apps_deleted_unit
\end{lstlisting}

\subsection{Outputs from VerityMath}

\subsubsection{Correct outputs} \label{correct UCPs}
\begin{lstlisting}[basicstyle=\scriptsize, escapechar={|}]
def solution():
    """Kelly has 5 quarters and 2 dimes. If she buys a can of pop for 55 cents, how many cents will she have left?"""
    quarters = 5
    quarters_unit = Counter({"quarters": 1})
    dimes = 2
    dimes_unit = Counter({"dimes": 1})
    cents_per_quarter = 25
    cents_per_quarter_unit = Counter({"cents": 1, "quarters": -1})
    cents_per_dime = 10
    cents_per_dime_unit = Counter({"cents": 1, "dimes": -1})
    total_cents_quarters = quarters * cents_per_quarter
    total_cents_quarters_unit = Counter({"cents": 1})
    assert total_cents_quarters_unit == quarters_unit + cents_per_quarter_unit
    total_cents_dimes = dimes * cents_per_dime
    total_cents_dimes_unit = Counter({"cents": 1})
    assert total_cents_dimes_unit == dimes_unit + cents_per_dime_unit
    total_cents = total_cents_quarters + total_cents_dimes
    total_cents_unit = Counter({"cents": 1})
    assert total_cents_unit == total_cents_quarters_unit == total_cents_dimes_unit
    cents_needed_for_pop = 55
    cents_needed_for_pop_unit = Counter({"cents": 1})
    cents_left = total_cents - cents_needed_for_pop
    cents_left_unit = Counter({"cents": 1})
    assert cents_left_unit == total_cents_unit == cents_needed_for_pop_unit
    result = cents_left
    return result
    
=> Example 287 in the test dataset. 

def solution():
    """Every day, Wendi feeds each of her chickens three cups of mixed chicken feed, containing seeds, mealworms and vegetables to help keep them healthy. She gives the chickens their feed in three separate meals. In the morning, she gives her flock of chickens 15 cups of feed. In the afternoon, she gives her chickens another 25 cups of feed. How many cups of feed does she need to give her chickens in the final meal of the day if the size of Wendi's flock is 20 chickens?"""
    cups_of_feed_per_chicken = 3
    cups_of_feed_per_chicken_unit = Counter({"cups": 1, "chickens": -1})
    chickens_flock_size = 20
    chickens_flock_size_unit = Counter({"chickens": 1})
    total_cups_of_feed_first_meal = 15
    total_cups_of_feed_first_meal_unit = Counter({"cups": 1})
    total_cups_of_feed_second_meal = 25
    total_cups_of_feed_second_meal_unit = Counter({"cups": 1})
    total_cups_of_feed_given = total_cups_of_feed_first_meal + total_cups_of_feed_second_meal
    total_cups_of_feed_given_unit = Counter({"cups": 1})
    assert total_cups_of_feed_given_unit == total_cups_of_feed_first_meal_unit == total_cups_of_feed_second_meal_unit
    remaining_cups_of_feed = (chickens_flock_size * cups_of_feed_per_chicken) - total_cups_of_feed_given
    remaining_cups_of_feed_unit = Counter({"cups": 1})
    assert remaining_cups_of_feed_unit == chickens_flock_size_unit + cups_of_feed_per_chicken_unit
    result = remaining_cups_of_feed
    return result

=> Example 5 in test dataset.

def solution():
    """Tom's ship can travel at 10 miles per hour. He is sailing from 1 to 4 PM. He then travels back at a rate of 6 mph. How long does it take him to get back?"""
    travel_rate_forward = 10
    travel_rate_forward_unit = Counter({"miles": 1, "hours": -1})
    time_forward = 4 - 1
    time_forward_unit = Counter({"hours": 1})
    distance_forward = travel_rate_forward * time_forward
    distance_forward_unit = Counter({"miles": 1})
    assert distance_forward_unit == travel_rate_forward_unit + time_forward_unit
    travel_rate_backward = 6
    travel_rate_backward_unit = Counter({"miles": 1, "hours": -1})
    time_backward = distance_forward / travel_rate_backward
    time_backward_unit = Counter({"hours": 1})
    assert time_backward_unit == distance_forward_unit - travel_rate_backward_unit
    result = time_backward
    return result

=> Example 52 in test dataset.

\end{lstlisting}

\subsubsection{Incorrect outputs} \label{incorrect UCPs}
The challenge of UCPs lies in the requirement to generate accurate \counter{} and \assert{} statements. Any program with incorrect \counter{} or \assert{} statements is immediately deemed incorrect due to the assertion errors that are raised, leading to decreased performance.
\begin{lstlisting}[basicstyle=\scriptsize, escapechar={|}]
def solution():
    """Two cups of flour are needed to make a dozen cookies. Carla is making 36 cookies today and 30 cookies tomorrow. How many cups of flour will Carla need to bake the cookies today and tomorrow?"""
    cups_per_dozen_cookies = 2
    cups_per_dozen_cookies_unit = |\color{red}Counter({"cups": 1, "dozen": -1})|
    cookies_today = 36
    cookies_today_unit = Counter({"cookies": 1})
    cookies_tomorrow = 30
    cookies_tomorrow_unit = Counter({"cookies": 1})
    total_cookies = cookies_today + cookies_tomorrow
    total_cookies_unit = Counter({"cookies": 1})
    assert total_cookies_unit == cookies_today_unit == cookies_tomorrow_unit
    dozens_of_cookies = total_cookies / 12
    dozens_of_cookies_unit = Counter({"dozens": 1})
    assert dozens_of_cookies_unit == total_cookies_unit - Counter({"cookies": 1, "dozens": -1})
    total_cups_needed = dozens_of_cookies * cups_per_dozen_cookies
    total_cups_needed_unit = Counter({"cups": 1})
    assert total_cups_needed_unit == dozens_of_cookies_unit + cups_per_dozen_cookies_unit
    result = total_cups_needed
    return result

=> Example 261 in test dataset. In this case the variable cups_per_dozen_cookies_unit should have a Counter({"cups": 1, "dozens": -1}).

def solution():
    """Ashley has an internet connection speed of 20kb per second. Knowing that 1 Mb has 1000 kb, she wants to know her internet connection speed in MB per hour. What is Ashley's internet connection speed in Mb per hour?"""
    speed_kb_per_second = 20
    speed_kb_per_second_unit = Counter({"kb": 1, "seconds": -1})
    kb_per_second_to_kb_per_hour = 3600 
    kb_per_second_to_kb_per_hour_unit = |\color{red}Counter({"kb": 1, "seconds": -1, "hours": -1})|
    speed_kb_per_hour = speed_kb_per_second * kb_per_second_to_kb_per_hour
    speed_kb_per_hour_unit = Counter({"kb": 1, "hours": -1})
    assert speed_kb_per_hour_unit == speed_kb_per_second_unit + kb_per_second_to_kb_per_hour_unit
    kb_per_mb = 1000
    kb_per_mb_unit = Counter({"kb": 1, "mb": -1})
    speed_mb_per_hour = speed_kb_per_hour / kb_per_mb
    speed_mb_per_hour_unit = Counter({"mb": 1, "hours": -1})
    assert speed_mb_per_hour_unit == speed_kb_per_hour_unit - kb_per_mb_unit
    result = speed_mb_per_hour
    return result

=> Example 627 in test dataset. In this case the variable kb_per_second_to_kb_per_hour_unit should have a Counter({"seconds": 1, "hours": -1}).

def solution():
    """Milo is making a mosaic with chips of glass. It takes twelve glass chips to make every square inch of the mosaic. A bag of glass chips holds 72 chips. Milo wants his mosaic to be three inches tall. If he has two bags of glass chips, how many inches long can he make his mosaic?"""
    chips_per_square_inch = 12
    chips_per_square_inch_unit = |\color{red}Counter({"chips": 1, "square inches": -1})|
    chips_per_bag = 72
    chips_per_bag_unit = Counter({"chips": 1, "bags": -1})
    bag_count = 2
    bag_count_unit = Counter({"bags": 1})
    total_chips = chips_per_bag * bag_count
    total_chips_unit = Counter({"chips": 1})
    assert total_chips_unit == chips_per_bag_unit + bag_count_unit
    total_square_inches = total_chips / chips_per_square_inch
    total_square_inches_unit = |\color{red}Counter({"square inches": 1})|
    assert total_square_inches_unit == total_chips_unit - chips_per_square_inch_unit
    mosaic_height_inches = 3
    mosaic_height_inches_unit = Counter({"inches": 1})
    mosaic_width_inches = total_square_inches / mosaic_height_inches
    mosaic_width_inches_unit = Counter({"inches": 1})
    assert mosaic_width_inches_unit == total_square_inches_unit - mosaic_height_inches_unit
    result = mosaic_width_inches
    return result

=> Example 808 in the test dataset. In this case the variable chips_per_square_inch_unit should have a Counter({"chips": 1, "inches": -2}) and total_square_inches_unit should have a Counter({"inches": 2}).

\end{lstlisting}

\end{document}